%% file: main.tex
\documentclass[twoside]{article}

\usepackage[accepted]{aistats2020}
\usepackage[round]{natbib}

\usepackage{hyperref}       
\usepackage{url}            
\usepackage{booktabs}       
\usepackage{amsfonts}       
\usepackage{nicefrac}       
\usepackage{microtype}      
\usepackage[english]{babel}
\usepackage{graphicx}
\usepackage{framed}
\usepackage[normalem]{ulem}
\usepackage{amsmath}
\usepackage{amsthm}
\usepackage{amssymb}
\usepackage{amsfonts}
\usepackage{enumerate}
\usepackage{color}
\usepackage{algorithm}
\usepackage{algpseudocode}
\usepackage{subfig}
\usepackage{comment}

\theoremstyle{definition}
\newtheorem{theorem}{Theorem}
\newtheorem{lemma}{Lemma}

\newcommand{\indep}{\rotatebox[origin=c]{90}{$\models$}}
\newcommand{\nindep}{\not\!\perp\!\!\!\perp}

\newcommand{\vz}{\mathbf{z}}

\begin{document}


\twocolumn[

\aistatstitle{Differentiable Causal Backdoor Discovery}

\aistatsauthor{ Limor Gultchin \And Matt J. Kusner \And  Varun Kanade \And Ricardo Silva}

\aistatsaddress{ University of Oxford \\ The Alan Turing Institute \And  University College London \\ The Alan Turing Institute \And University of Oxford \\ The Alan Turing Institute \And University College London \\ The Alan Turing Institute} ]





\begin{abstract}
\input{sections/abstract.tex}
\end{abstract}

\section{Introduction}
\input{sections/introduction.tex}

\section{Background}
\input{sections/background.tex}

\section{Method}
\input{sections/methods.tex}
\section{Experiments}
\input{sections/experiments.tex}

\section*{Acknowledgements}

This work was supported by the Alan Turing Institute under the EPSRC grant EP/N510129/1.

\bibliography{bibliography,rbas}
\bibliographystyle{icml2019}


  %
  %

  \appendix
  \renewcommand{\thesubsection}{\thesection.\arabic{subsection}}

  \newcommand{\beginsupplementary}{%
    \setcounter{table}{0}
    \renewcommand{\thetable}{S\arabic{table}}%
    \setcounter{figure}{0}
    \renewcommand{\thefigure}{S\arabic{figure}}%
    \setcounter{theorem}{0}
    \renewcommand{\thetheorem}{\arabic{theorem}}%
    \setcounter{lemma}{0}
    \renewcommand{\thelemma}{\arabic{lemma}}%
    
  }
  \newcommand{\suptitl}{Supplementary Material for: \\ Differentiable Causal Backdoor Discovery}
  \beginsupplementary
  \onecolumn
  \aistatstitle{\suptitl}
  \runningtitle{}
  \setlength\headheight{0pt} 
  \setlength\headsep{0pt}
  
  \input sections/proofs

\end{document}

%% file: sections/abstract.tex
Discovering the causal effect of a decision is critical to nearly all forms of decision-making. In particular, it is a key quantity in drug development, in crafting government policy, and when implementing a real-world machine learning system. Given only observational data, confounders often obscure the true causal effect. Luckily, in some cases, it is possible to recover the causal effect by using certain observed variables to adjust for the effects of confounders. However, without access to the true causal model, finding this adjustment requires brute-force search. In this work, we present an algorithm that exploits auxiliary variables, similar to instruments, in order to find an appropriate adjustment by a gradient-based optimization method. We demonstrate that it outperforms practical alternatives in estimating the true causal effect, without knowledge of the full causal graph.

%% file: sections/introduction.tex
Causal modelling allows one to go beyond observational quantities 
to estimate the \emph{causal effect} of interventions in the real-world  \citep{pearl:00,rosenbaum:17}. 
Most commonly, we are interested in how a single outcome $Y$ varies as we change values of a decision or treatment $X$. However, without knowledge of the true causal graph, estimating the causal effect from observational data can be confounded by unobserved variables. While it is possible to ``adjust'' for these confounding variables, one needs to know what observed variables to use in the adjustment. Otherwise, the adjusted estimate can be worse than the unadjusted estimate \citep{pearl:09a}.

\begin{figure}[t!]
    \centering
    \includegraphics[width=0.8\columnwidth]{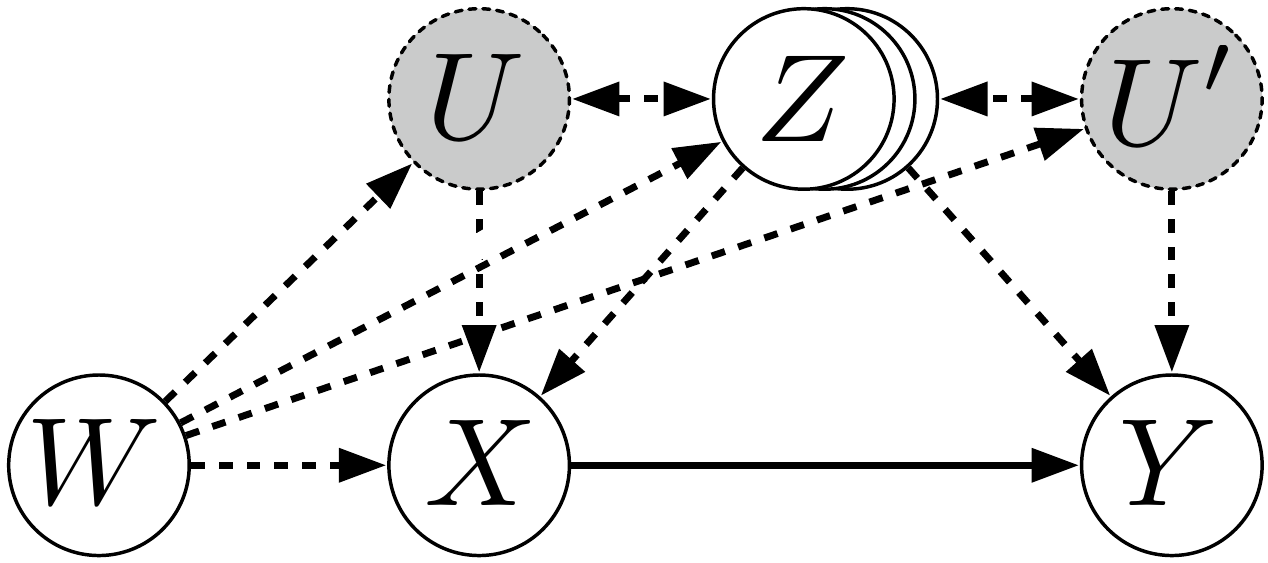}
    \vspace{-1ex}
    \caption{An informal visualization of our assumed knowledge of the causal graph. Treatment $X$ must be a parent of outcome $Y$, and dashed arrows/nodes are optional as long as a subset $Z^\star$ of $Z$ blocks the backdoors between $X$ and $Y$, and $W$ has no unblocked path into $Y$ given $X$ and $Z^\star$. Additionally, $W$ must be associated with $X$ (and possibly confounded).}
    \vspace{-3ex}
    \label{fig:setup}
\end{figure}

One way to solve this is to use causal discovery algorithms \citep{sgs:00,peters:17} to identify as much of the causal graph as possible, and then make an adjustment. However, there are a number of problems:  
1. In general, causal discovery algorithms involve expensive combinatorial optimization problems; 
2. Traditional nonparametric causal discovery methods \citep{sgs:00} do not discover the full causal graph, but a Markov-equivalent set of graphs that imply the same set of conditional independences. This is an issue for adjustment as edge existence and directionality play key roles in whether a variable should be used in an adjustment or not;
3. Nonparametric methods also require combinations of multiple tests of independence constraints which suffer from rapidly diminishing power as dimensionality grows.



Our contribution is to provide a continuous optimization approach to the problem of learning what variables should be used for adjustment. Although the optimization is non-convex and our practical implementation makes parametric assumptions, our search problem has important benefits compared to causal discovery methods \citep{sgs:00,peters:17}. Specifically, it avoids all of the problems listed above as follows: 
1. Instead of resorting to combinatorial search, our method uses gradient-based optimization that runs extremely fast on modern hardware to discover the adjustment set;
2. Our method does not need to know the full connectivity of the causal graph. All that is needed is to identify a suitable auxiliary variable $W$ and covariate set $Z$ which precede the treatment $X$ and outcome $Y$;
3. Our method directly targets functions of the covariate space $Z$ that are useful for covariate adjustment. Thus it does not directly perform high-dimensional multiple-testing and does not assume a small number of parents for the variables of interest.

Our goal is to provide a generally-applicable algorithm to be added to the toolbox of the practitioner, while focusing on the ``causal supervised learning'' problem of targeting a given cause-effect pair $(X, Y)$ as opposed to the full graph learning problem \citep{zheng:18}. We will begin by reviewing causal primitives necessary for our work in the next section.


%% file: sections/background.tex

Using the notation of \citet{pearl:00}, we let the quantity $P(Y \!=\! y~|~do(X \!=\! x), Z \!=\! \vz)$ describe the
variability of $Y$ under an intervention that fixes $X$ at value $x$, conditioned on observations $\vz$ of a set of random variables $Z$ (throughout: non-bold lower-case variables $x$ are scalars, bold lower-case $\vz$ are vectors, non-bold capital $X$ are either single random variables, or a set of random variables in the case of $Z$). 
This is different from the usual conditional probability of $Y$ given $\{X \!=\! x, Z \!=\! \vz\}$, denoted, $P(Y \!=\! y~|~X \!=\! x, Z \!=\! \vz)$. 

We can formalize interventional probabilities using a 
directed acyclic graph (DAG) in the following way. Given a DAG
$\mathcal G$ with vertex set $\{V_1, V_2, \dots, V_d\}$, let
$\mathrm{pa}_{\mathcal G}(i)$ be the parents of $V_i$ in $\mathcal
G$. When each vertex corresponds to a random variable, this DAG induces a probabilistic model. In this model the probability (density or mass) function over
$\{V_1, \dots, V_d\}$ factorizes as $\Pi_{i = 1}^d
p(v_i~|~\mathrm{pa}_{\mathcal G}(i))$ \citep{lauritzen:96}. Each edge into $V_i$ in the graph qualitatively describes a contribution to model
factor $p(v_i~|~\mathrm{pa}_{\mathcal G}(i))$. 
Given the DAG and associated model, the intervention
$do(V_A =
v_A)$ corresponds to a two-step procedure: (a) remove all model factors
$p(v_A~|~\mathrm{pa}_{\mathcal G}(A))$ (i.e., all edges entering $V_A$ in the graph), and (b) fix the value of $V_A$ to $v_A$ in any other factor where $V_A$ is a parent node. 
We say that a model is
\emph{causal} if the \emph{do-operator} $do(\cdot)$ is defined for it. If so, then $\mathcal
G$ is a \emph{causal graph}, and $V_i$ \emph{causes} $V_j$ only if
$V_i$ is an ancestor of $V_j$ in $\mathcal G$.

One key summary of the interventional distribution $P(Y~|~do(X \!=\! x))$ is
the \emph{average treatment effect} (ATE),
\begin{align}
\mathbb E[Y~|~do(X = x')]
- \mathbb E[Y~|~do(X = x)], \nonumber
\end{align}
defined for two treatment levels $x$ and $x'$. Another effect of interest is the partial derivative
$\partial \mathbb E[Y~|~do(X = x)]/\partial x$, which we will also refer to as the ATE whenever $X$ is continuous. This will be our main
case in the sequel. The classic way to estimate ATE is via 
a randomized control trial.  However, with  observational data only, all we can directly estimate is the joint
distribution and an adjustment will be necessary based on the causal graph.  In particular, if there are common causes of $X$
and $Y$ (\emph{confounders}), then off-the-shelf regression of $Y$ on
$X$ can be severely biased.

To fix this, one approach is \emph{covariate adjustment}: find a set $Z^\star \subseteq Z$ of
ancestors of $X$ or $Y$ in $\mathcal G$ that can ``block'' such common causes and
apply a formula such as the \emph{backdoor adjustment} \citep{pearl:00},
\begin{equation}
  p(y~|~do(X = x)) = \int_{\vz^\star} p(y~|~x, \vz^\star)p(\vz^\star)d\mathbf z^\star. \label{eq:back}
\end{equation}
We say that $Z^\star$ is a \emph{valid} covariate set if it satisfies the above.
Notice that the marginalization is with respect $p(\mathbf z^\star)$ instead of
$p(\mathbf z^\star~|~x)$, as the link between $X$ and its ancestors in $\mathcal G$
is broken by the $do$ operator.

\paragraph{Finding valid covariate adjustments.}

If the full graph is known, there is a graphical criterion by which we
can test whether $Z^\star$ is a valid set for covariate adjustment \citep[cf.][for details]{pearl:00}. However, specifying a full causal graph is often difficult, 
particularly when all we need is to provide
a valid covariate set for a given cause-effect pair $(X, Y)$. As formalized by~\cite{ilya:11}, partial knowledge of the causal structure may suffice.

\begin{figure}[t!]
    \centering
    \includegraphics[width=\columnwidth]{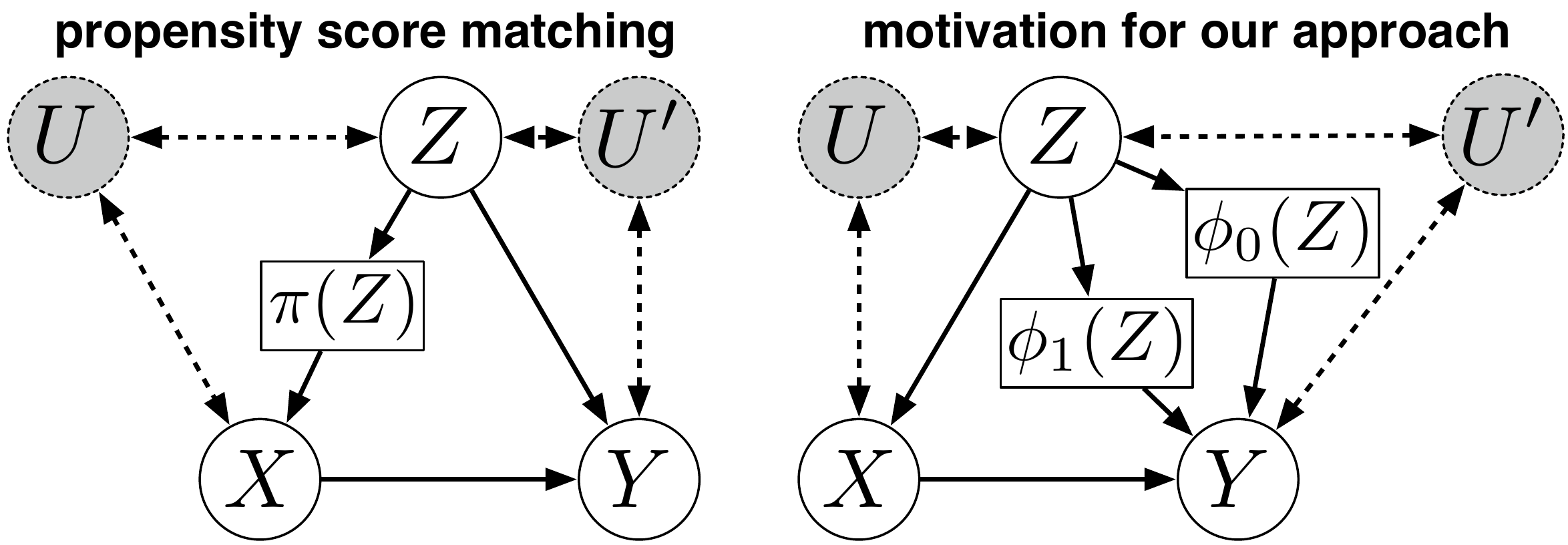}
    \vspace{-4ex}
    \caption{On the left, a representation of the propensity score $\pi(z) \equiv p(x~|~z)$ as a vertex that satisfies the backdoor criterion once placed in the graph. On the right, an analogous representation, in terms of two extra vertices, of the outcome functions $P(Y = 1~|~z, x)$ for $x = 0$ ($\phi_0$) and $x = 1$ ($\phi_1$). Here, the value taken by $X$ works as a selection indicator of a mixture model so that, for binary $Y$, $p(y~|~x, z) = \phi_x(z)^y(1 - \phi_x(z))^{1 - y}$. In fact, $\phi_x(z)$ can be the result of any invertible transformation of $P(Y = 1~|~x, z)$.}
    \vspace{-2ex}
    \label{fig:inspiration}
\end{figure}

Consider the causal setup in Figure~\ref{fig:setup} (solid arrows indicate a causal link, dashed arrows indicate a causal link may or may not exist). Let $Z \cup \{W\}$ be a known set of observed non-descendants of $\{X, Y\}$, and $U,U'$ possible unobserved parents of $Z \cup \{X\}$ and $Z \cup \{Y\}$. Assuming faithfulness \citep{sgs:00}, \citet{entner:13} observed that it is possible to recover the causal effect $X \rightarrow Y$ between treatment $X$ and outcome $Y$ so long as an observed adjustment set $Z^\star \subseteq Z$ satisfies the following criterion for some observed variable $W$:
\begin{align}
    W &\,\indep \;Y~|~Z^\star \cup \{X\}, \label{eq:entner} \\
    W &\nindep Y~|~Z^\star. \nonumber
\end{align}
That is, $Z^\star$ is a valid covariate set ``certified'' by an auxiliary variable $W$. In a simplified sense, $W$ plays the role of a pseudo ``intervention indicator'' into $X$ with all paths from $W$ into $Y$ mediated by $X$. We choose to focus in this work on the difficult step on finding a valid $Z^{\star}$. An auxiliary variable $W$ can be found in linear time by an outer loop, or by choosing based on background knowledge (which still requires much weaker conditions than full graph elicitation). In fact, different criteria can be used to combine multiple candidate $W$s \citep{silva2016causal}. Our contribution therefore is in identifying a valid covariate adjustment set, and we assume the existence of a suitable W in the following.
The benefit of the above criteria is that it much weaker partial ordering assumptions as opposed to knowing a full graph. While there are many covariate adjustment strategies \citep[see][for a recent review]{witte2019covariate}, we are unaware of any that use similarly weak or weaker assumptions.


\paragraph{Our work.}

Finding a valid covariate set satisfying these criteria in general requires
combinatorial optimization on $Z$.  This is usually done by greedy/random search \citep{entner:13}. In this paper, we propose instead a fully-differentiable optimization problem for learning a backdoor adjustment. Instead of attempting to find the exact adjustment set
our approach 
finds a set of functions 
\[
\Phi(\mathbf z) \equiv \{\phi_x(\mathbf z)~|~x \in \mathcal X\},
\]
where $\mathcal X$ is the sample space of $X$ and $\mathbf z$ is in the sample space of $\mathbf Z$, such that
\begin{align}
    W &\,\indep \;Y~|~\{\phi_X(Z^\star), X\}, \label{eq:phi_entner}\\
    W &\nindep Y~|~\Phi(Z^\star). \nonumber
\end{align}
We will show that, under some general conditions, covariates $Z^\star$ satisfying eq.~\eqref{eq:phi_entner} will also satisfy eq.~\eqref{eq:entner}. Importantly, to simplify the presentation, we will assume that either $Y$ is binary or the whole causal system is linear Gaussian. This will allow us to define $\phi_x(\mathbf z)$ to be scalars for any $\mathbf z$\footnote{More precisely, we assume we can write $p(y~|~\mathbf z', x) \equiv h(y, x, \phi_x(\mathbf z'))$ for some $h(\cdot)$ and any subvector $\mathbf z'$ of $\mathbf z$.}. The extension to non-binary $Y$ or non-linear Gaussian systems is conceptually simple, but notation gets considerably more evolved.

For intuition about why this is true, first consider a graphical representation of the propensity score, $\pi(\mathbf z) \equiv \mathrm{Pr}(X = 1~|~\mathbf z)$ added to a postulated causal graph $\{Z \rightarrow X, Z \rightarrow Y, X \rightarrow Y\}$ and binary $X$, This is shown in Figure \ref{fig:inspiration}: here informally $Z$ is a set of vertices, where single vertex $\pi(Z)$ \citep[see][Chapter 15, for a more formal discussion]{hernan:20} blocks all backdoors between $X$ and $Y$ and hence is a valid adjustment variable. This is particularly helpful as a way of reducing the dimensionality of the problem, if we can reasonably estimate $\pi(Z)$. However, discovering this backdoor adjustment by finding a suitable $Z^\star$ will not be possible if, for instance, $W$ and $Z^\star$ are adjacent in the causal graph: we will still need to explicitly condition on $Z^\star$ when verifying the independence between $W$ and $Y$. 

An alternative is to consider the analogue to the propensity score with respect to the outcome variable $Y$, as illustrated  by the following example.

\paragraph{Example 1.} Assume all variables are binary, with $\log[\mathrm{Pr}(Y \!=\! 1~|~X, Z) / (1 - \mathrm{Pr}(Y \!=\! 1~|~X, Z))] = (1 - X)\beta_{yz0}^\top Z + X\beta_{yz1}^\top Z$ and $\log[\mathrm{Pr}(X \!=\! 1~|~W, Z) / (1 - \mathrm{Pr}(X \!=\! 1~|~W, Z))] = (1 - W)\beta_{xz0}^\top Z +
W\beta_{xz1}^\top Z$, with $W$ being an exogenous variable. Then we can check that $\Phi(Z) \equiv \{\phi_0(Z) \equiv \beta_{yz0}^\top Z, \phi_1(Z) \equiv \beta_{yz1}^\top Z\}$ will satisfy $W \indep Y~|~\{\phi_X(Z), X\}$ and $W \nindep Y~|~\Phi(Z)$. The former can be shown by noting that we can predict $Y$ purely from $X$ and $\phi_X(Z)$; no further information about $W$ will help. The latter can be verified by observing that $W$ provides further information about $X$, which we can use to refine our prediction of $Y$. A graphical illustration of this idea is shown in Figure \ref{fig:inspiration}. $\Box$

This suggests that if we parameterize $\phi_x(\mathbf z)$ to be in the same family of the response of $Y$ given $X$ and any other set of observable covariates, we will be able to directly search for this representation without performing high-dimensional tests of conditional independence. However, this raises the immediate concern of what to do if $X$ is continuous, as in this case $\Phi(\mathbf z)$ is uncountable. We can compress the information in $\Phi(\mathbf z)$ by making further assumptions about the outcome regression model, as shown in the following example.

\paragraph{Example 2.} Assume that $Y = \beta_{yx} X + \beta_{yz}^\top Z + \epsilon_y$ and $X = \beta_{xw} W + \beta_{xz}^\top Z + \epsilon_x$ describe the conditional distributions of $Y$ and $X$, with $W$ being an exogenous variable and $\epsilon_x, \epsilon_y$ being independent error terms. Then we can check that $\phi_X(Z) \equiv \beta_{yz}^\top Z$ for all $X$ will satisfy $W \indep Y~|~\{\phi_X(Z), X\}$ and $W \nindep Y~|~\Phi(Z)$. $\Box$

We will prove the existence of a solution of eq.~\eqref{eq:phi_entner} that solves eq.~\eqref{eq:entner} in the following section. This suggests we can obtain a valid adjustment from the optimization of functions $\phi_X(\cdot)$. 
Instead of searching for exact conditional independence, 
our approach is to 
minimize dependence measures motivated by eq.~\eqref{eq:phi_entner}. 
In the parametric case, we will derive a continuous optimization problem, avoiding greedy/random selection \citep{entner:13}. While continuous optimization methods exist for discovering the entire causal graph \citep{mooij:09,zheng:18}, our technique is tailored to discovering a backdoor adjustment.


It is tempting to see our definition of $\phi_X(Z)$ as a similar idea to
propensity scores \citep{hernan:20}, as the examples suggest: can we make $Y$ and $Z$ independent given $X$ and $\phi_X(Z)$, so that covariate adjustment can be done directly with $\phi_X(Z^\star)$ as opposed to $Z^\star$? Unfortunately, this is not true: eq.~\eqref{eq:phi_entner} can only identify an equivalence class of $\phi_X(Z)$, not all of which will be a valid adjustment by itself. For instance, in the first example, regressing $W$ using $X$ and $Z$ will also satisfy eq.~(\ref{eq:phi_entner}). Therefore, when we solve for $\phi_X(Z)$, our goal is to discover which variables $Z^\star \subseteq Z$ should make up its domain. In the next section, we give conditions that allow us to identify a suitable adjustment set $Z^\star$ via $\phi_X(\cdot)$, and we describe how to solve for it. 




%% file: sections/methods.tex
We now describe the general problem formulation, starting with the idealized
scenario where we know the true population observational distribution. Practical
implementations of this formulation for linear models with homoscedastic errors are discussed in the sequel.

Let $d(V_i, V_j~|~S)$ be a measure of probabilistic dependence between random variables $V_i$ and $V_j$ given a set of random variables $S$. Let $d(\cdot, \cdot~|~\cdot)$ have the following properties: (a) it is non-negative, and (b) it equals zero if and only if $V_i \indep V_j~|~S$ \footnote{If $d$ is a probabilistic dependence measure, (a) and (b) are necessary and sufficient conditions to define a valid optimization problem as described in \eqref{eq:problem}.}. An example of such a measure is the conditional mutual information. In linear models, absolute partial correlation could be used. Let $\phi_X(\vz)$ have a parametric representation, with $\theta_X$ being the respective
parameters. Let $\mathrm{sparsity}(\{\theta_X\})$ be a penalty term that induces sparsity in this set of parameter vectors parameter vector $\theta$, e.g. $\sum_x\Vert \theta_x \Vert_1$. 
We define the following optimization problem for $\{\theta_X\}$:
\begin{equation}
\begin{array}{rl}
  \mathrm{minimize} & 
  d(W, Y~|~X, \phi_X(Z))   \\ 
 \mathrm{subject~ to}       & d(W, Y~|~\Phi(Z))  > \alpha, \\
									 & \mathrm{sparsity}(\{\theta_X\}) < c.
\end{array}
\label{eq:problem}
\end{equation}



\subsection{Theory Behind Learning $\phi_X$}

In this section, we will present the theoretical justification of our method. The main idea is to show the following: i) if $W \indep Y~|~Z^\star \cup \{X\}$ for some $Z^\star$, then there exists some scalar $\phi_X(Z^\star)$ where $W \indep Y~|~\{\phi_X(Z^\star), X\}$; ii) if $W \indep Y~|~\{\phi_X(Z^\star), X\}$,  then $W \indep Y~|~\{X\} \cup  Z^\star$ up to some ``general'' arrangement of the parameters of the model; iii) under \emph{faithfulness} (conditional independences in the data arise from conditional independences in the causal graph), we can search for a $Z^\star$ satisfying ii), and use it to estimate the ATE using the backdoor adjustment with adjustment set $Z^\star$. All results assume the partial ordering described in Figure 1, and that for simplicity of presentation $Y$ is binary or the system is linear-Gaussian so that each $\phi_x(\mathbf z)$ can be written as scalar. All proofs are in the supplement.

Point i) was implicitly discussed in the previous section, and it is formalized here for the general case where some elements of $Z^\star$ are not parents of $Y$: 

\begin{lemma}
If $W \indep Y~|~Z^\star \cup \{X\}$, then there exists some scalar $\phi_X(Z^\star)$ such that $W \indep Y~|~\{\phi_X( Z^\star), X\}$.
\end{lemma}

The result for point ii) is as follows. To simplify the presentation,  we assume that $Z$ follows a multivariate discrete distribution, but this is not essential. We also define $\Phi_{x \mathbf{z}^\star}^f$ to be subset of the sample space of $ Z^\star$ such that $\phi_x(\mathbf z^\star) \!=\! f$ for all $\mathbf z^\star \!\in\! \Phi_{x\mathbf{z}^\star}^f$. We assume the causal model is parametric with a Lebesgue measure on the parameter space.  As an abuse of notation, we sometimes use $p(x_i, X_j = f, \dots)$ to mean the pmf $p(x_i, f, \dots)$ of the joint distribution of $X_i, X_j, \dots$, when it is not obvious that ``$f$'' is a value taken by $X_j$.

\begin{theorem}
  If $W \nindep Y~|~Z^\star \cup \{X\}$, and
  
  \[
    \displaystyle
    \sum_{\mathbf{z}^\star \in \Phi_{x\mathbf{z}^\star}^f}p(y~|~w, x, \mathbf{z}^\star)\frac{p(\mathbf{z}^\star~|~w, x)}{\mathrm{Pr}(Z^\star \in \Phi_{x\mathbf{z}^\star}^f~|~w, x)} \neq
  \]
  \begin{equation}    
    \sum_{\mathbf{z}^\star \in \Phi_{x\mathbf{z}^\star}^f}p(y~|~x, \mathbf{z}^\star)\frac{p(\mathbf{z}^\star~|~x)}
    {\mathrm{Pr}(Z^\star \in \Phi_{x\mathbf{z}^\star}^f~|~x)},
  \label{eq:lemma1}
  \end{equation}

  \noindent for some value $f$ in the range of $\phi_x(\cdot)$, then $W \nindep Y~|~\{\phi_X( Z^*), X\}$.
\label{theorem:wy_ind}
\end{theorem}


One observation: equation (\ref{eq:lemma1}) is a relationship that we do not expect to hold in
general, even if it can hold for particular parameter arrangements. For instance, if for simplicity $Z \indep W~|~X$, then the last step of the derivation above can be written
as
\[
  \displaystyle
  \sum_{\mathbf{z}^\star \in \Phi_{x\mathbf{z}^\star}^f}p(y~|~w, x, \mathbf{z}^\star)\frac{p(\mathbf{z}^\star~|~x)}{P(Z^\star\in \Phi_{x\mathbf{z}^\star}^f~|~x)} =
\]
\[
  \displaystyle
  \sum_{\mathbf{z}^\star \in \Phi_{x\mathbf{z}^\star}^f}p(y~|~x, \mathbf{z}^\star)\frac{p(\mathbf{z}^\star~|~x)}{P(Z^\star \in \Phi_{x\mathbf{z}^\star}^f~|~x)},  
\]
\noindent this is $\mathbb{E}[p(y~|~w, x, Z^\star) \!-\! p(y~|~x, Z^\star)~|~X \!=\! x, \phi_x(Z^\star) \!=\! f] = 0$. This is not
an identity 
if $p(y~|~w, x, \mathbf{z}^\star) \neq p(y~|~x, \mathbf{z}^\star)$. In this
sense, we say that ``in general'' $W \nindep Y~|~\{X, Z^\star\}$ implies $W \nindep Y~|~\{X, \phi_X(Z^\star)\}$ so that the contrapositive holds.

The main exception to avoid is the case where $\phi_x(\mathbf z^\star)$ is functionally independent of some elements of $Z^\star$. A simple example is the graph $\{W \rightarrow Z_1 \leftarrow U \rightarrow Y, W \rightarrow X, X \rightarrow Y\}$: inequality (\ref{eq:lemma1}) \emph{will} be violated for $\phi_x(z_1) \!=\! \mathrm{constant}$ under \emph{any} parameterization, even though we expect $W \nindep Y | \{Z_1, X\}$. Hence, it is crucial to return functions $\phi_X(\cdot)$ that are as sparse as possible.

Our method then follows from using the optimization problem in eq.~(\ref{eq:problem}) to find a function $\phi_x(\mathbf z^\star)$ for each $x$, that is \emph{sparse}. Specifically, it may be a function of a strict subset of $Z$. Assuming (a) that the conditions of Theorem \ref{theorem:wy_ind} hold; (b) that our choice of $W$, variable selection algorithm, and function space for $\phi_X(\cdot)$ can minimize $d(W, Y~|~X, \phi_X(Z))$ all the way to zero; (c) $W$ is a parent of $X$ so that the second condition of \citet{entner:13} is satisfied ($W \nindep Y~|~\{X\} \cup Z^\star$), then by faithfulness we recover a valid adjustment set for the causal effect of $X$ and $Y$. 

Given these results, we will show in the next section how to optimize a fully-differentiable objective 
for covariate adjustment search that does not require solving a full graph discovery problem. The price to be paid is the assumption about the existence of an auxiliary variable $W$. This assumption is nevertheless falsifiable by simply testing whether we can indeed minimize our objective function to zero. Note that technically we need to prove similar results for the second term in eq.~\eqref{eq:phi_entner} ($W \nindep Y~|~\phi_X(Z^\star)$) to show that we do not need the edge $W \rightarrow X$. 
We omit those proofs for simplicity of presentation, as they are nearly identical. 
We now describe an optimization procedure to find a differentiable backdoor adjustment $\phi_X(Z)$.

\subsection{Optimization \& Implementation }\label{sec:optimization}

The shape of the function $\phi_X(Z)$ should be decided based on the assumptions about the outcome model $p(y~|~x, \vz^\star)$. We will consider Gaussian models. In particular, consider the following,
\begin{align*}
    X &\sim \mathcal{N}(g_\textrm{con}(Z, W, U), \sigma_X^2) \\
    Y &\sim \mathcal{N}(h_\textrm{con}(X, Z, U), \sigma_Y^2), 
\end{align*}
where $g_\textrm{con}(\cdot, \cdot, \cdot), h_\textrm{con}(\cdot, \cdot, \cdot) \in \mathbb{R}$ are unknown functions. One possible modelling assumption for $g_\textrm{con}(\cdot, \cdot, \cdot), h_\textrm{con}(\cdot, \cdot, \cdot)$ which we consider here is that they are linear functions of their inputs. Specifically, this implies that: $p(y~|~x, \vz) = p(y~|~\alpha x + \boldsymbol{\beta}^\top \vz)$ for some $(\alpha, \beta)$. Thus, we could set $\phi_X(\vz) \equiv \phi(\vz) \equiv \boldsymbol{\beta}^\top \vz$. In this case we use absolute partial correlation $|\rho(\cdot, \cdot | \cdot)|$ as our probabilistic dependence measure as follows
\begin{align}
    d(W, Y~|~X, \phi(Z)) = 
    | \rho(W, Y~| X, \phi(Z)) |.
     \nonumber
\end{align}
To learn $\boldsymbol{\beta} \in \mathbb{R}^d$, the parameters of $\phi(Z)$, an initial idea would be to solve for it directly by minimizing the absolute partial correlation. However, recall that the partial correlation is computed from a ratio of terms. In our case, both the numerator and denominator include $\boldsymbol{\beta}$. This makes the partial correlation scale invariant with respect to $\boldsymbol{\beta}$. Thus, we reparameterize $\boldsymbol{\beta}$ as $\boldsymbol{\beta} \equiv \boldsymbol{\gamma} / \Vert \boldsymbol{\gamma} \Vert_2$ and propose to optimize the Lagrange equivalent of eq.~\eqref{eq:problem}, which is:
\begin{align}
\min_{\boldsymbol{\gamma}}&
\left| \rho(W, Y~|X, \boldsymbol{\beta}^\top Z) \right| - \lambda_1\left|\rho(W, Y~| \boldsymbol{\beta}^\top Z ) \right| + \lambda_2 \Vert \boldsymbol{\beta}  \Vert_1, \nonumber \\
\mbox{s.t.}&\; \boldsymbol{\beta} \equiv \boldsymbol{\gamma} / \Vert \boldsymbol{\gamma} \Vert_2. \nonumber
\end{align}

Although the above objective is non-convex, it is differentiable everywhere except $\boldsymbol{\beta}\!=\!0$ and hence amenable to optimization using gradient-based methods. In the next section we demonstrate our method on simulated graphs and graphs fit on real-world data compared to other practical baselines.

%% file: sections/experiments.tex
To evaluate our method we begin by devising a high-dimensional simulation benchmark. This benchmark will allow us (a) to test the robustness of our technique to different causal graph parameter settings, (b) to study the sensitivity of our method to different noise and causal effect parameters. 
Additionally, we devise a causal graph based on real-world health-worker survey data. We compare our method with practical baselines including \citet{entner:13}. In all cases, our method matches or outperforms all baselines. Code to replicate experiments and run on new data will be released at \url{https://github.com/limorigu/Diff-causal-backdoor-disc}.

\begin{figure*}[t!]
\vspace*{-5ex}
    \centering
    \subfloat[absolute ATE error for each method on datasets with lower noise on treatment ($\sigma_X^2\!=\!0.01$).]{{\includegraphics[scale=0.3]{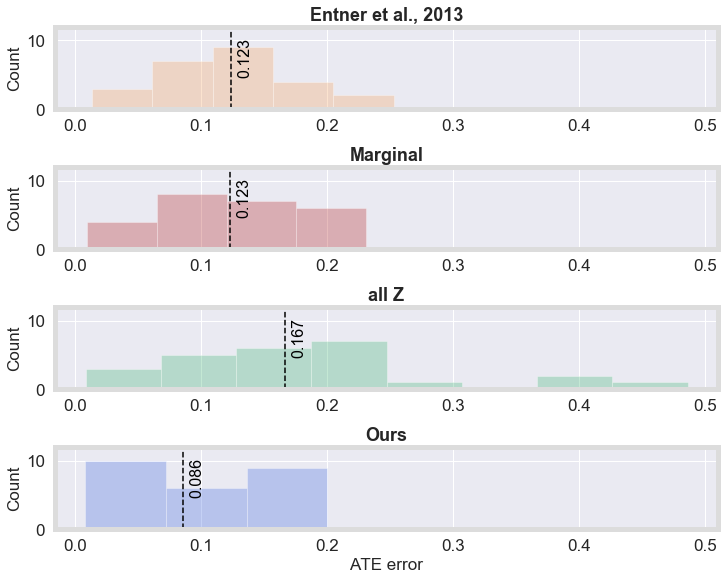}}}%
    \qquad
    \subfloat[absolute ATE error for each method on datasets with higher noise on treatment ($\sigma_X^2\!=\!0.6$).]{{\includegraphics[scale=0.3]{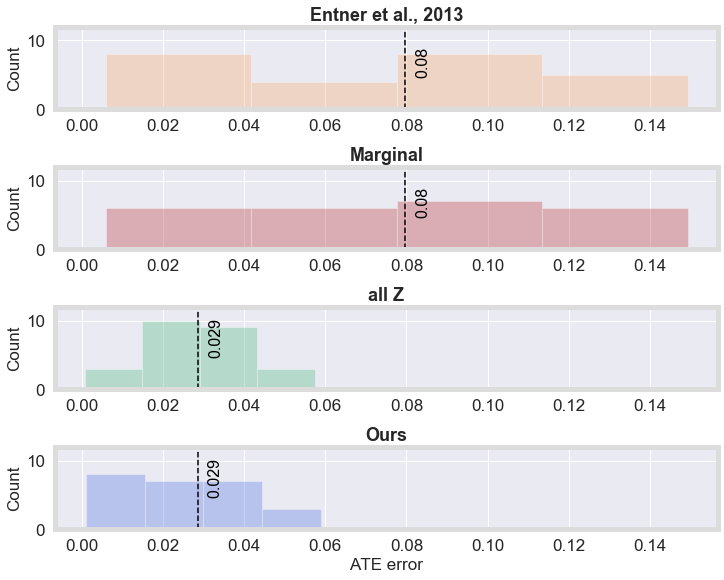} }}%
    \vspace{-1.5ex}
    \caption{Histograms of ATE error of all methods on a simulation with lower treatment effect ($\omega\!=\!0.1$).}
    \label{fig:hist_allZ0.1}%
    \vspace*{-3ex}
\end{figure*}


\begin{figure*}[t!]%
    \centering
    \subfloat[ATE error of baselines vs our method for simulations with lower noise on treatment ($\sigma_X^2\!=\!0.01$).]{{\includegraphics[scale=0.23]{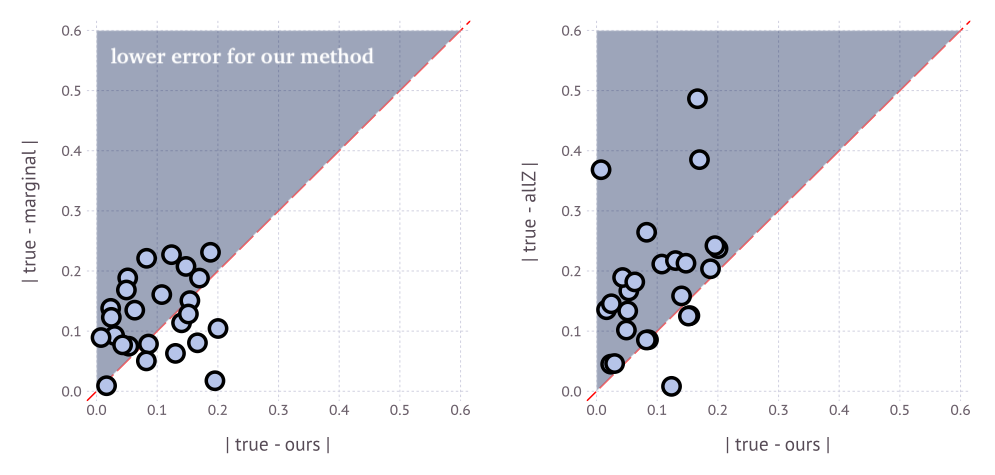}}}%
    \qquad
    \subfloat[ATE error of baselines vs. our method for simulations with higher noise on treatment ($\sigma_X^2\!=\!0.6$).]{{\includegraphics[scale=0.23]{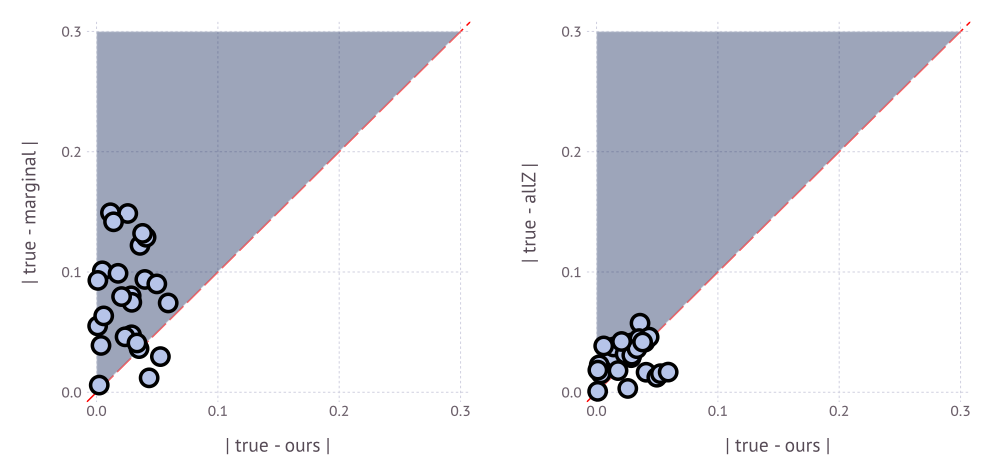}}}%
    \vspace{-0.5ex}
    \caption{Scatter plot comparing baselines and our method on a simulation with lower treatment effect ($\omega\!=\!0.1$). Each point is one of the 25 parameter settings. Points in blue regions indicate our method performs better.}
    \label{fig:scatter_xeffect0.1}%
    \vspace{-2ex}
\end{figure*}


\begin{figure}[t!]
    \centering
    \includegraphics[scale=0.4]{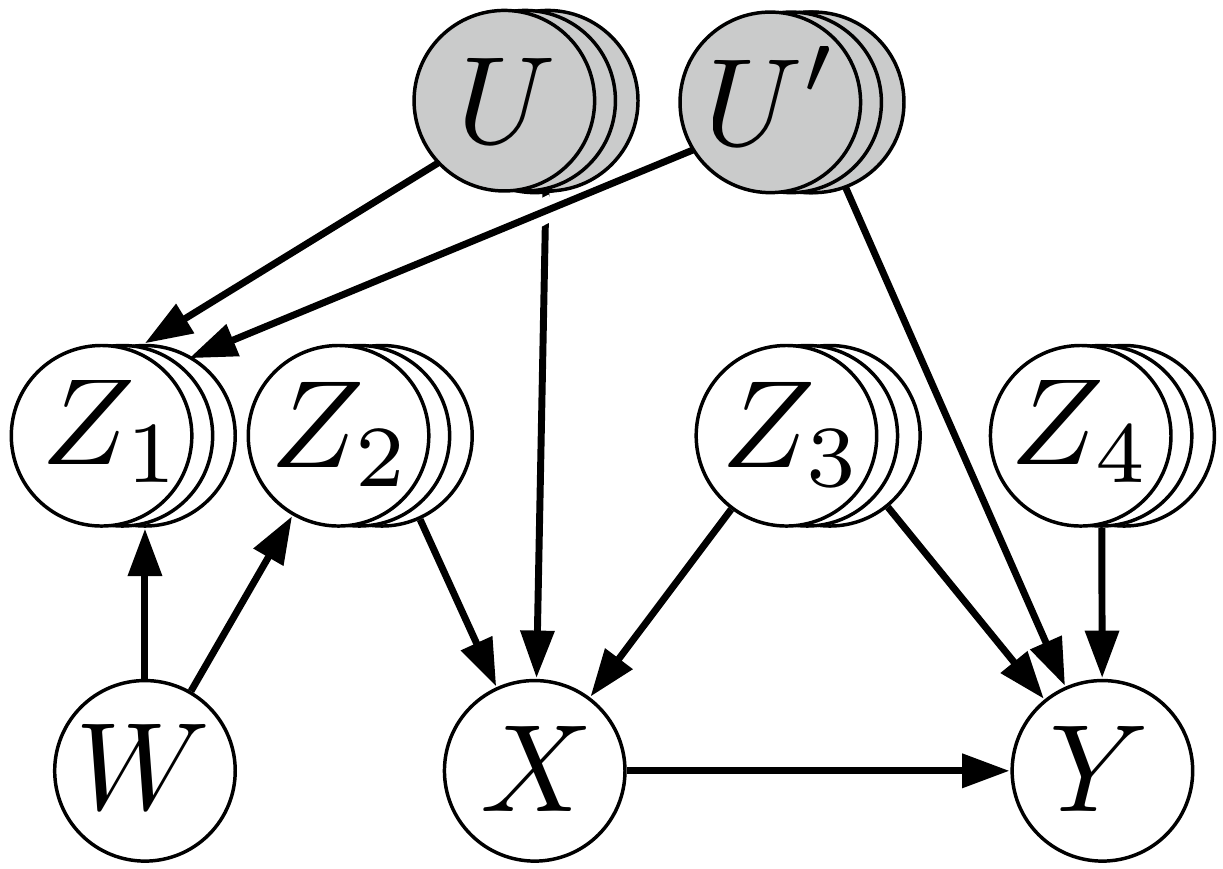}
    \caption{The causal graph used in our simulation. There are four (unknown) covariate sets $Z_1, Z_2, Z_3, Z_4$, two unobserved variable sets $U,U'$, an auxiliary variable $W$, treatment $X$, and outcome $Y$. Notice that the minimal backdoor adjustment set is $Z_3$, while adjusting for $Z_1$ or $Z_2$  can adversely affect the estimation of the average treatment effect (ATE) of $X$ on $Y$.}
    \vspace{-3ex}
    \label{fig:sim}
\end{figure}

\subsection{Simulation Benchmark} \label{sec:simulation}

We design a simulation benchmark to test the robustness and sensitivity of our method. The causal graph is shown in Figure~\ref{fig:sim}. The covariates $Z$ are (secretly) grouped into four sets $Z_1, Z_2, Z_3, Z_4$. While adjusting for $Z_3$ allows us to correctly estimate the ATE of $X$ on $Y$, adjusting for $Z_1$ or $Z_2$ skews this estimate. We observe all variables except $U,U'$ and assume we have identified $W,X,Y,Z$ but do not know their connectivity except that it satisfies the partial ordering of
Figure~\ref{fig:setup}. Finally, as described in Section~\ref{sec:optimization} we will assume that the structural equations are linear Gaussian, which allows us to compare directly to \cite{entner:13}. 

The structural equations for the simulation are: 
\begin{equation}
\begin{aligned}[c]
        U \sim& \mathcal{N}(0,\mathbf{I}), U' \sim \mathcal{N}(0,\mathbf{I}), W \sim \mathcal{N}(0,1),\\
        Z_1 \sim&\; \theta_{Z_1, W} W + \boldsymbol{\Theta}_{Z_1, U} U + \boldsymbol{\Theta}_{Z_1, U'} U' + \mathcal{N}(0, \sigma_{Z_1}^2\mathbf{I}), \\
        Z_2 \sim&\; \theta_{Z_2, W} W + \mathcal{N}(0,\mathbf{I}), Z_3 \sim \mathcal{N}(0, \sigma_{Z_3}^2\mathbf{I}), Z_4 \sim \mathcal{N}(0,\mathbf{I}),\\
        X \sim&\; \boldsymbol{\theta}_{X,U}^\top U + \boldsymbol{\theta}_{X,Z_2}^\top Z_2 + \boldsymbol{\theta}_{X,Z_3}^\top Z_3 + \mathcal{N}(0, \sigma_X^2), \\
        Y \sim&\; \boldsymbol{\theta}_{Y, U'}^\top U' + \boldsymbol{\theta}_{Y,Z_2}^\top Z_2 + \boldsymbol{\theta}_{Y,Z_4}^\top Z_4 + \omega X + \mathcal{N}(0, \sigma_{Y}^2)
\end{aligned}
\nonumber
\end{equation}
where $\Theta$ signify matrices of parameters. Our goal is to learn $\phi(Z)$ in order to correctly estimate the ATE: $\omega$.

\begin{figure*}[t!]
\vspace*{-5ex}
    \centering
    \subfloat[absolute ATE error for each method/baseline on datasets with lower noise on treatment ($\sigma_X^2\!=\!0.01$).]{{\includegraphics[scale=0.3]{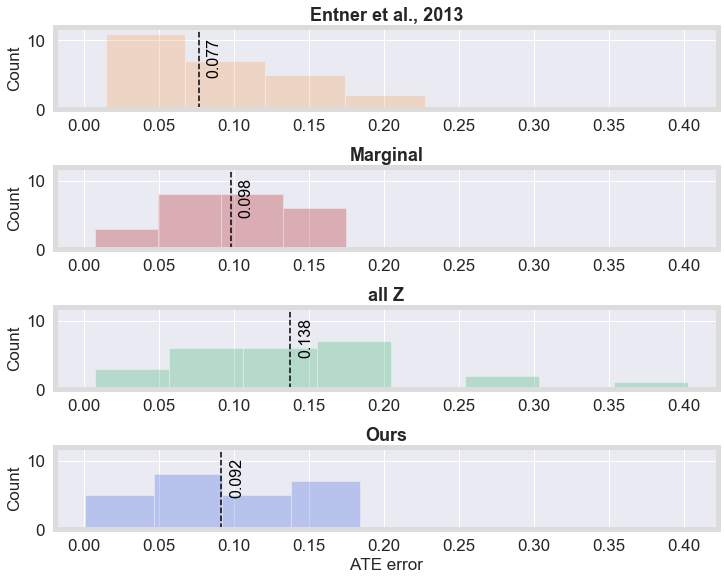} }}%
    \qquad
    \subfloat[absolute ATE error for each method/baseline on datasets with higher noise on treatment ($\sigma_X^2\!=\!0.6$).]{{\includegraphics[scale=0.3]{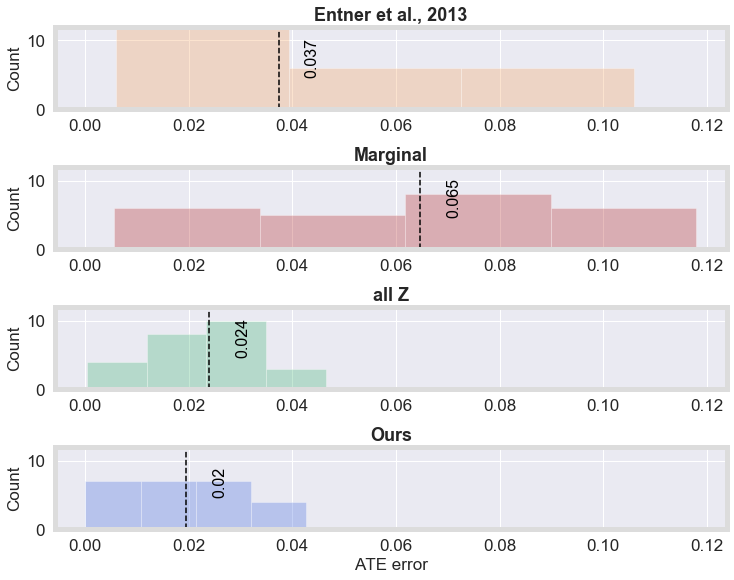} }}%
    \vspace*{-1.5ex}
    \caption{Our method's performance compared to the marginal, all Z and \citeauthor{entner:13}'s method baselines on a simulation with higher treatment effect ($\omega\!=\!0.5$).}%
    \label{fig:hist_xeffectpoint5}
    \vspace*{-2ex}
\end{figure*}


\begin{figure*}[t!]%
    \centering
    \subfloat[Comparison for simulation with lower noise on treatment ($\sigma_X^2\!=\!0.01$).]{{\includegraphics[scale=0.23]{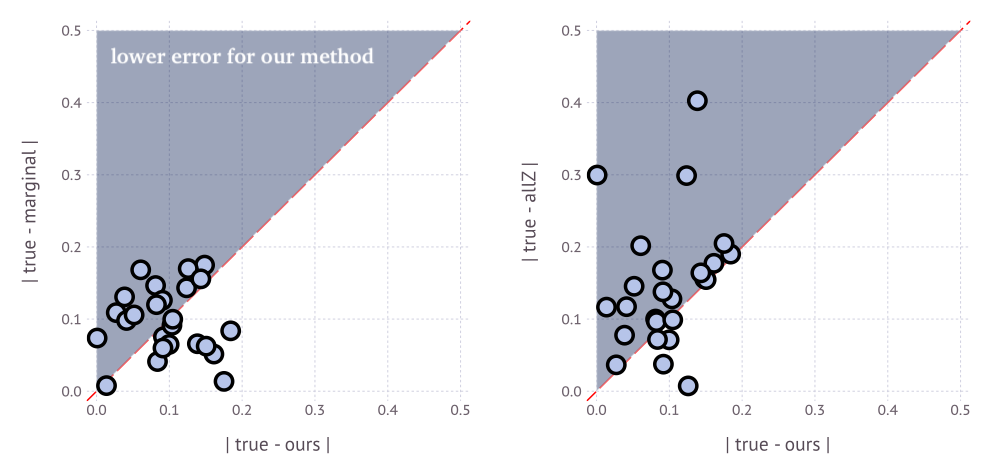}}}%
    \qquad
    \subfloat[Comparison for simulation with higher noise on treatment ($\sigma_X^2\!=\!0.6$).]{{\includegraphics[scale=0.23]{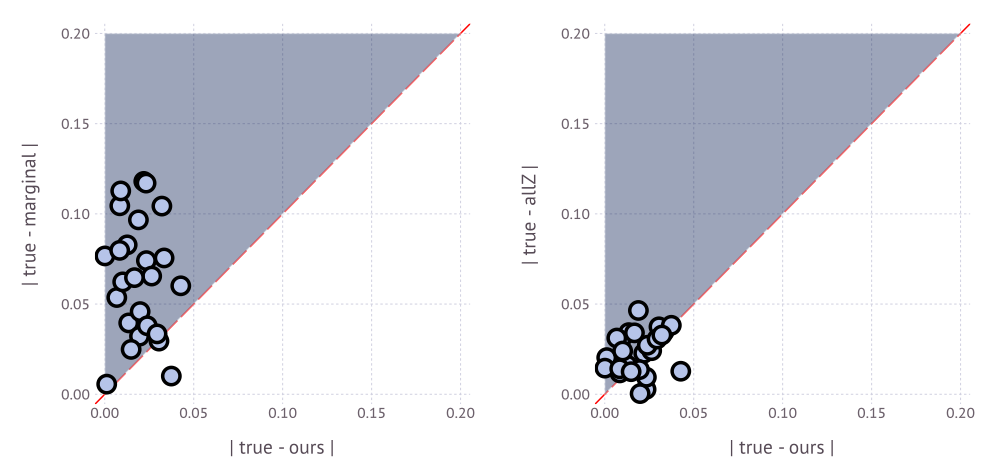}}}%
    \vspace*{-1.3ex}
    \caption{Performance of our method against the marginal and the allZ baslines on a simulation with a higher effect of treatment ($\omega\!=\!0.5$). Scatter plots represent 25 different datasets.}%
    \label{fig:scatter_xeffect0.5}%
    \vspace*{-2ex}
\end{figure*}

\paragraph{Baselines.}
We compare our approach with a set of practical baselines that make similar assumptions:
\begin{enumerate}
\item \citep{entner:13}: We compare against the ATE estimated by the high-dimensional search algorithm of \citet{entner:13}. 
\item \textbf{all Z}: This uses all covariates $Z$ in the adjustment set to compute the ATE.
\item \textbf{marginal}: This uses no covariates in the adjustment set to compute the ATE.
\end{enumerate}
As described in Section 2, there exist many algorithms for covariate selection \citep{witte2019covariate}. However none of them work under the weak assumptions of Figure 1, apart from \citet{entner:13}. Below, we simulate problems in which either baseline 2 or 3 perform well. Our experiments illustrate that our method is competitive even with baseline 1 that is tailored for linear Gaussian models, while matching or improving upon the best baseline. 
Thus, without knowledge of the true causal model, our model is state-of-the-art.
\textbf{Experimental setup.}
To evaluate the robustness of all methods we sample 25 different parameter settings for the above structural equations (we fix the true ATE $\omega$ throughout, more on this in the following paragraph). We sample parameters from a standard Gaussian distribution then take the absolute value. We then generate signs for parameters by drawing a value from a uniform random variable and flipping the current sign if  the value is above a certain threshold. This ensures that sampled parameters are not mean zero and thus will not likely cancel each other out (i.e., violating faithfulness). For each setting we then sample 20,000 inputs and normalize each variable to have unit variance. We split these inputs 50/50 into train/test. We fix the dimension of each variable $Z1, Z2, Z3, Z4, U, U'$ to $30$.

\begin{figure}[t!]
    \centering
    \includegraphics[width=0.8\columnwidth]{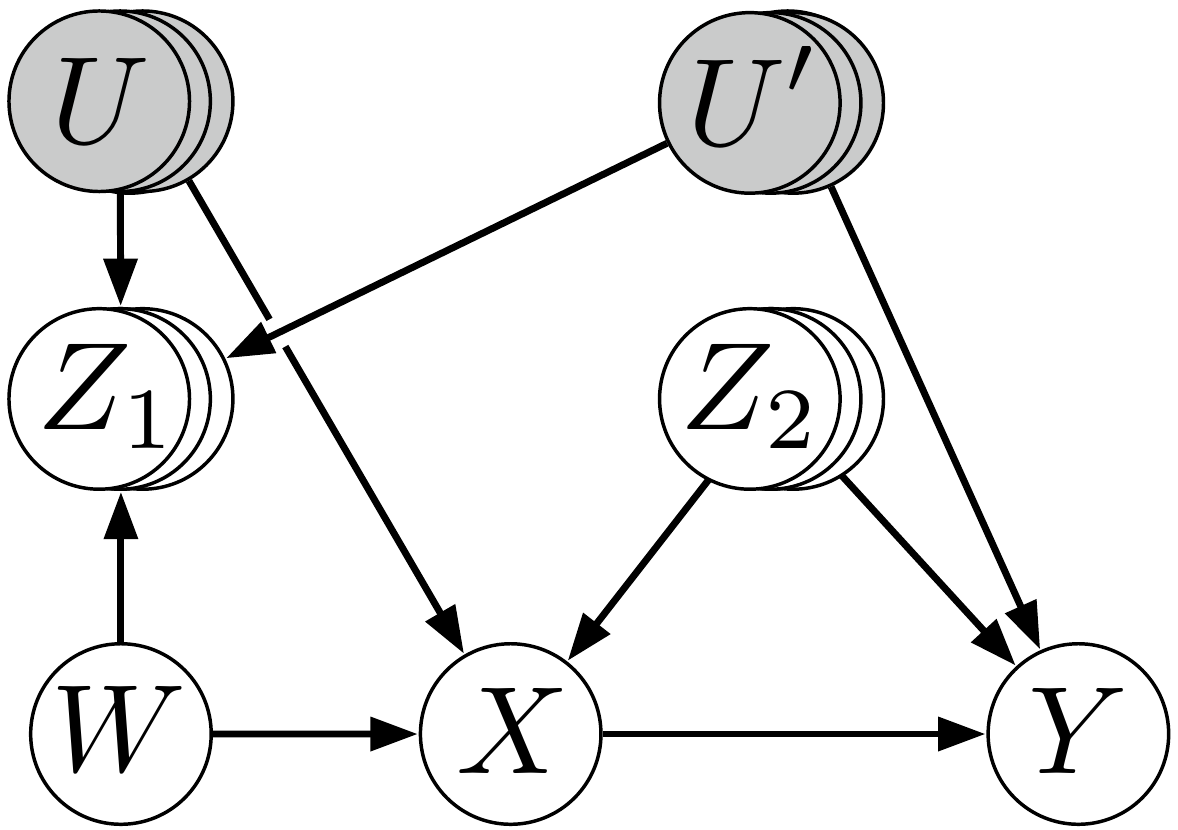}
    \caption{DAG representing the NHS dataset we use to test performance of our model.}
    \vspace{-3ex}
    \label{fig:real_world_DAG}
\end{figure}

\textbf{Hyperparameter tuning.} We tune all hyperparameters of our method including $\lambda_1, \lambda_2$, the initialization of $\boldsymbol{\gamma}$, and the learning rate $\eta$ on the training set. We use cross validation to select $\lambda_1$, initial $\boldsymbol{\gamma}$ and $\eta$. To select $\lambda_2$, we perform a hypothesis test with null $\rho(W, Y | \boldsymbol{\beta}^\top Z) \!=\! 0$. If we reject the null, we increase $\lambda_2$ and re-optimize (adjusting significance using Bonferroni correction) until we do not reject the null.

\textbf{Results.} To test the sensitivity of our method we introduced two types of problems: i) problems with high noise on treatment $X$ ($\sigma_X^2\!=\!0.6$), and ii) problems with lower noise on treatment $X$ ($\sigma_X^2\!=\!0.01$). Problem i) will cause the marginal baseline to perform poorly, while problem ii) will cause the all Z baseline to perform poorly. 
Additionally, for each problem we tested two settings of the true ATE: $\omega=0.1$ and $\omega=0.5$. 
Thus, we have 2 problem types, each with 2 different ATEs, tested over 25 parameter settings, for a total of 100 different simulation scenarios.

Figures~\ref{fig:hist_allZ0.1} and \ref{fig:scatter_xeffect0.1} show test results on the low ATE setting ($\omega\!=\!0.1$) in two different ways. Figure~\ref{fig:hist_allZ0.1} shows a histogram of absolute ATE error for each method in both treatment noise settings $(\sigma_X^2\!=\!0.01)$ and $(\sigma_X^2\!=\!0.6)$. The black dashed line indicates the median of the distributions. In both cases our method matches or outperforms all other methods as measured by the median performance. Figure~\ref{fig:scatter_xeffect0.1} shows the distribution of performance for each individual parameter setting, compared with the marginal and all Z baselines. Points in the blue regions indicate trials where our method outperformed the baseline methods. In the low noise plots (a) our method consistently outperforms the all Z method, while in the high noise plots (b) our method noticeably outperforms the marginal method.

Figures~\ref{fig:hist_xeffectpoint5} and \ref{fig:scatter_xeffect0.5} present the same type of results for the setting with higher effect of treatment X ($\omega\!=\!0.5$). In this setting the high-dimensional method of \citet{entner:13} deteriorates. As in the previous settings,
our method does better than two baselines, and at least as well as the third.  This means that when one encounters an unknown dataset for which we want to estimate the ATE, one doesn't need to guess whether adjustment using all, none, or some other set of covariates is best. By applying our method one can get accurate ATE estimates regardless of the setting.

\subsection{NHS Health Data}
Alongside the simulation benchmark, we test our method on a causal graph derived from real-world health data. Specifically, we consider data from the 2014 UK National Health Service (NHS) Survey \cite{nhs2014staffsurvey}. The aim of the survey was to ``gather information that will help to improve the working lives of staff in the NHS''. We consider the goal of trying to understand the causal effect of workplace training on personal well-being.

We construct a set of variables by averaging the results of related questions (where most questions on a five-level Likert scale: from `strongly disagree' to `strongly agree'). Specifically, we identified an auxiliary variable $W$: whether an individual underwent workplace training (Q1 in the survey), a treatment $X$: one's benefit (or not) from training (Q2), outcome $Y$: whether an individual's job is good for their well-being (Q14), and covariates $Z1$: a set of variables describing personal job satisfaction (Q5-Q9), and $Z2$: a set describing effectiveness of one's organization/managers (Q10-Q12, Q18-Q21). Based on their descriptions we describe the relationships between these variables using the causal structure in Figure~\ref{fig:real_world_DAG}. Additionally, we include unobserved variables: $U$ could be interpreted as one's "openness/ability to learn" while $U'$ could be one's "personal affinity for their job". We use linear Gaussian structural equations for the model, and fit the parameters of the model using the real data, $25$ variables in total. Once fit, we sample 90,000 data-points and partition them into 30,000 train/valid/test splits. We ran each method on sampled data-points
(and tuned hyperparameters on the validation set) and evaluated them on the test set.

\textbf{Results.} 
Table~\ref{tab:real_world_results} shows the ATE error of each method on the NHS health data. Our method outperforms all other methods.
It is worth noting that for this dataset, the all Z baseline is much closer to the real ATE than the marginal baseline. Compared to the all Z baseline, our method removes
a harmful node in $Z_1$, which leads to a better ATE estimate. 

\begin{table}[t!]
\centering
\begin{tabular}{@{}cccc@{}}
\toprule
          Ours & All Z & Marginal & \citet{entner:13} \\ \midrule
          0.163 & 0.324 & 1.747 & 0.771 \\ \bottomrule
\end{tabular}%
\caption{The absolute ATE error for all methods on the NHS health data.}
\label{tab:real_world_results}
\vspace{-4ex}
\end{table}

\section{Conclusion}

In the spirit of \cite{entner:13}, we showed how causal discovery for covariate adjustment can be tackled directly without the need for full causal graph discovery. In the spirit of \cite{zheng:18} and \cite{mooij:09} we exploited how we can formulate the problem directly as a continuous optimization problem without the need of combinatorial search or indirectly optimizing a likelihood function.
To do so, we derived (in)dependence criteria conditional on functions of covariates. These criteria are sufficient for backdoor adjustment. By learning these functions to minimize dependence scores on the observed variables, we have a differentiable way to learn a backdoor adjustment that can control for unobserved confounders. We showed how our method consistently matched or outperformed baselines that make the same weak assumptions as our work.
There are many exciting directions for future work, including formulating semi-parametric versions of the method and considering problems where $W$ and $X$ may be sets of instruments/treatments. In this case, stronger signals may be obtained, making the problem more realistic in practice if the goal is to properly control for favourable outcomes of $Y$.

%% file: sections/proofs.tex
\section{Proofs}
\label{sec:proofs}

\begin{lemma}
If $W \indep Y~|~Z^\star \cup \{X\}$, then there exists some scalar $\phi_X(Z^\star)$ such that $W \indep Y~|~\{\phi_X( Z^\star), X\}$.
\end{lemma}

\paragraph{Proof.} We will discuss the case for binary $Y$, as the proof for linear-Gaussian models follows the same idea. Let the structural equation for $Y$ be given by $f_y(x, \pi_Z, \pi_U)$, where $\pi_Z$ and $\pi_U$ are the observed and unobserved parents of $Y$ in the corresponding causal graph. The conditional distribution of $Y$ is given by
\begin{align}
p(y~|~x, \mathbf z^\star) = \nonumber\\
p(f_y(x, \pi_Z, \pi_U) = 1~|~x, \mathbf z^\star)^y \times \nonumber\\
(1 - p(f_y(x, \pi_Z, \pi_U) = 1~|~x, \mathbf z^\star))^{1 - y}.
\nonumber
\end{align}
By assumption, $f_y(\cdot)$ is functionally independent of $W$. Now we just have to show that the random variable $f_y(x, \pi_Z, \pi_U)$ is conditionally independent of $W$ given $X$ and $Z^\star$. Since $W \indep Y~|~Z^\star \cup \{X\}$, it cannot be the case that $W$ and $\pi_{\backslash Z^\star, X}$, the parents of $Y$ not in $Z^\star \cup \{X\}$, are conditionally dependent given $Z^\star \cup \{X\}$. We define $\phi_X(\mathbf z^\star)$ as $p(f_y(x, \pi_Z, \pi_U) = 1~|~x, \mathbf z^\star)$ for each possible realization of $X$. Given $X$, we can fully reconstruct from $\phi_X(\mathbf z^\star)$ a conditional distribution of $Y$ that makes information about $W$ irrelevant. $\Box$

\begin{theorem}
  If $W \nindep Y~|~Z^\star \cup \{X\}$, and
  
  \[
    \displaystyle
    \sum_{\mathbf{z}^\star \in \Phi_{x\mathbf{z}^\star}^f}p(y~|~w, x, \mathbf{z}^\star)\frac{p(\mathbf{z}^\star~|~w, x)}{\mathrm{Pr}(Z^\star \in \Phi_{x\mathbf{z}^\star}^f~|~w, x)} \neq
  \]
  \begin{equation}    
    \sum_{\mathbf{z}^\star \in \Phi_{x\mathbf{z}^\star}^f}p(y~|~x, \mathbf{z}^\star)\frac{p(\mathbf{z}^\star~|~x)}
    {\mathrm{Pr}(Z^\star \in \Phi_{x\mathbf{z}^\star}^f~|~x)},
  \label{eq:lemma1}
  \end{equation}

  \noindent for some value $f$ in the range of $\phi_x(\cdot)$, then $W \nindep Y~|~\{\phi_X( Z^*), X\}$.
\nonumber
\end{theorem}

\paragraph{Proof.} 
Assume, contrary to the hypothesis, that $W \indep Y~|~\{ \phi_X(Z^\star), X\}$. Then
\[
  p(y~|~w, x, \phi_x(Z^\star) = f) = p(y~|~x, \phi_x(Z^\star) = f) \Rightarrow
\]
\[
  \sum_{\mathbf{z}^\star} p(y~|~w, x, \phi_x(Z^\star) = f, \mathbf z^*)p(\mathbf{z}^*~|~w, x, \phi_x(Z^\star) = f) = 
\]
\[
  \sum_{\mathbf{z}^\star} p(y~|~x, \phi_x(Z^\star) = f, \mathbf z^\star)p(\mathbf{z}^\star~|~x, \phi_x(Z^\star) = f) \Rightarrow
\]
\[
  \sum_{\mathbf{z}^\star} p(y~|~w, x, \mathbf{z}^\star)p(\mathbf{z}^\star~|~w, x, \phi_x(Z^\star) = f) = 
\]
\[
  \sum_{\mathbf{z}^\star} p(y~|~x, \mathbf{z}^\star)p(\mathbf{z}^*~|~x, \phi_x(Z^\star) = f) \Rightarrow
\]
\[
 \displaystyle
 \sum_{\mathbf{z}^\star \in \Phi_{x\mathbf{z}^\star}^f}p(y~|~w, x, \mathbf{z}^\star)\frac{p(\mathbf{z}^\star~|~w, x)}
 {\mathrm{Pr}(Z^\star \in \Phi_{x\mathbf{z}^\star}^f~|~w, x)} =
\]
\[
 \displaystyle
 \sum_{\mathbf{z}^\star \in \Phi_{x\mathbf{z}^\star}^f}p(y~|~x, \mathbf{z}^\star)\frac{p(\mathbf{z}^\star~|~x)}
  {\mathrm{Pr}(Z^\star \in \Phi_{x\mathbf{z}^\star}^f~|~x)},  
\]
which contradicts the hypothesis. $\hfill\Box$